\documentclass{llncs}

\usepackage{amssymb}
   \usepackage[neveradjust]{paralist}
  \usepackage{soul, color}
  \sethlcolor{green}
\usepackage{url}
\usepackage[pdftex]{graphicx}

 \usepackage{multicol}
  \usepackage{multirow}

\newcommand{\cqcnl}{{\sf CLaRO}}

\definecolor{mk}{RGB}{50, 50, 250}

\definecolor{rt}{RGB}{150, 150, 50}

\begin{document}

\title{\cqcnl: a Data-driven CNL for Specifying Competency Questions}

\author{C. Maria Keet, Zola Mahlaza, and Mary-Jane Antia}

\institute{Department of Computer Science, University of Cape Town, South Africa
            \email{\{mkeet,zmahlaza,mjantia\}@cs.uct.ac.za}}
            
\maketitle              

\begin{abstract}
Competency Questions (CQs) for an ontology and similar artefacts aim to provide insights into the contents of an ontology and to demarcate its scope. 
The absence of a controlled natural language, tooling and automation to support the authoring of CQs has hampered their effective use in ontology development and evaluation.
The few question templates that exists are based on informal analyses of a small number of CQs and have limited coverage of question types and sentence constructions. 
We aim to fill this gap by proposing a template-based CNL  to author CQs, called \cqcnl. For its design, we exploited a new dataset of 234 CQs that had been processed automatically into 106 patterns, which we analysed and used to design a template-based CNL, with an additional CNL model and XML serialisation. 
The CNL was evaluated with a subset of questions from the original dataset and with two sets of newly sourced CQs. The coverage of \cqcnl, with its 93 main templates and 41 linguistic variants, is about 90\% for unseen questions. \cqcnl~has the potential to facilitate streamlining formalising ontology content requirements and, given that about one third of the competency questions in the test sets turned out to be invalid questions, assist in writing good questions.  
\end{abstract}

\section{Introduction}

The specification of Competency Questions (CQ) is step in the process of the development of ontologies and similar artefacts---called ``OMS'' in \cite{JYB-Festschrift2015-DOL}, for Ontologies, Models and Specifications. CQs aim to provide insights into the contents of an ontology, to demarcate its scope, and, ideally, be used in the verification step during testing of the model. They function alike requirements in the traditional requirements engineering setting, but then are formulated as questions that such an OMS should be able to answer. For instance, {\sf Do lions eat grass?} that some wildlife ontology may have to be able to answer, {\sf Which software can perform clustering?} for a structured controlled vocabulary about software, and {\sf What are the related terms of propaganda?} for the ERIC thesaurus. 
CQs have been emphasised over the years as a key requirement for ontology development \cite{Uschold96} and form part of, among others, the NeON methodology for ontology development \cite{Suarez08} and are an option in test-driven ontology development \cite{KL16}. 
However, CQs are rarely published at all or in full 
except in a few cases, notably, \cite{Malone14,DEMCARE}. 
Two main reasons put forward for their low uptake are, firstly, the lack of guidance for formalising them---be this in SPARQL, SPARQL-OWL, OWL or another language---which affects requirements testing of the ontology, and, secondly, the `free text' nature of CQs makes operationalising them to test an ontology against far from trivial.

A well-known solution direction to such problems is to constrain the natural language so as to streamline the input, which facilitate their formalisation into the desired target logic or query language. A few CQ types, patterns, and ``archetypes'' have been proposed based on a manual analysis of a small set of 
CQs \cite{Ren14,Bezerra14}, which goes in the direction of a controlled natural language (CNL). However, their 12 resp. 14 patterns are merged with types of ontology elements, therewith constraining its usage to OWL and a particular modelling style, and their adequacy, or coverage, is unknown. Currently, no CNL exists for CQs that has been shown to be adequate in coverage and be at the natural language layer.

In this paper, we seek to address these shortcomings by developing a CNL for CQs. We reuse the CQ dataset and analysis of \cite{WPLK18} that consists of 234 type-level CQs for five ontologies and the 106 data-driven CQ patterns based on them. 
Based on the analysis of the patterns and other design decisions, we convert those patterns into a template-based CNL, called \cqcnl: 
{\bf C}ompetency question {\bf La}nguage for specifying {\bf R}equirements for an {\bf O}ntology, model, or specification. \cqcnl~is evaluated against a random selection of CQs from the CQ dataset \cite{WPLK18} for verification and against a newly collected set of 20 CQs that were not part of the training set and half (21) of the Pizza CQs. \cqcnl's coverage was found to range from good to excellent and substantially outperforming the related work.
Overall, this resulted in 93 core templates and 41 variants, which cover about 90\% of the CQs of the test sets.

The remainder of this paper is structured as follows. Related work is discussed in Section~\ref{sec:relwork}. The CNL design and evaluation are described in Sections~\ref{sec:cnl} and~\ref{sec:eval}, respectively. We provide some examples of \cqcnl's use in Section~\ref{sec:ex}. We discuss in Section~\ref{sec:disc} and conclude in Section~\ref{sec:concl}.

\section{Related work}
\label{sec:relwork}
Over the years, CQs have been proposed for use in several areas such as education, in school-based initial teacher education \cite{Williams96}, law, assessment of the rights of mentally disabled individuals for or against the administration of certain treatments \cite{Perlin90}, improving the ability of maltreated children to provide needed answers under oath \cite{Lyon01}. In ontology engineering, CQs have been identified to play important roles in ontology development for demarcation of the scope of an ontology and alignment of source and target ontologies \cite{Uschold96,Thieblin18}, verification and evaluation \cite{Bezerra17,Azzaoui13,Bezerra:2013:EOC:2568492.2569217,KL16}. In spite of their acknowledged importance in ontology engineering, CQs are hardly available publicly, 
with the exception of the CQ sets for the Software Ontology (SWO) \cite{Malone14} and Dem@Care \cite{DEMCARE}. Wi\'sniewski et al. \cite{WPLK18} recently compiled 234 CQs from 5 ontologies into a freely available dataset\footnote{\url{https://github.com/CQ2SPARQLOWL/Dataset}}, including the SWO and Dem@Care CQ sets, and analysed the questions with NLP to chunk it and replace nouns and verbs with variables for entities and predicates, resulting in 106 CQ patterns, which are also part of that dataset. Those CQ patterns do not constitute a Controlled Natural Language (CNL), but may be useful input for specifying one. Related to this effort is earlier work by Ren and co-authors \cite{Ren14,Dennis17}, who  analysed about 150 CQs from two ontologies (SWO and Pizza) and proposed 12 core CQ ``archetypes'' and 7 variants, which also go in the direction of a CNL. In this case, however, those archetypes incorporate ontology elements explicitly, using 1:1 mappings between noun or noun phrase and OWL class (``[CE]'') and verb and OWL object property (``[OPE]''); e.g., ``Which [CE1] [OPE] [CE2]?''. 
Bezerra et al. \cite{Bezerra14} proposed 3 CQ types and 14 patterns for a template-based CNL for OWL ontologies, also based on a 1:1 mapping between natural language and ontology. 
Ren et a. has two patterns in incorrect English (``Be there...'') and Bezerra et al. has two templates missing a class variable after the initial start-of-sentence text and missing text in the two templates for disjointness, and three which/what confusions that is a recurring aspect in actual CQs as well\footnote{Note: `which' is used for choice or selection among a limited set of options, `what' does not have a count delimiter; e.g., ``which animals eats grass?' and `Which directed tasks are mono tasks?' is about a selection among a set of animals/tasks,  whereas `what kind of homogeneous mixture is mayonnaise?' and `What are the protocol parts?' are open-ended ``give me anything and all that satisfies''.}. 
They have limited coverage, however, such as no ``Who...'' or ``Where...'' questions in \cite{Bezerra14}, yet they do exist in the SWO CQs and are in Ren et al.'s archetypes, yet a simple subclass request, like {\sf DemCare\_CQ\_8.What are the types of diagnosis?} has no matching pattern in \cite{Ren14} but has in 
\cite{Bezerra14}, and, e.g., only the archetype {\em 1e.Be there [CE1] with [CE2]?} in \cite{Ren14}  but not also, say, `in' instead of `with' to accommodate the slightly different CQ, like {\sf awo9.Are there [these animals] in [this country]?} from the larger CQ set of \cite{WPLK18}. Conversely, there is only one negative question in \cite{WPLK18}'s set ({\sf awo\_5.Is there an animal that does not drink water?}) and one implicit disjointness ({\sf stuff\_04.Can a solution be a pure stuff?}), whereas \cite{Bezerra14,Ren14} have three templates with negation that are all motivated by the Pizza CQ set. 

Malheiros et al's approach with grammatical tags and regex rules for the remaining part of the sentence takes a step in the direction of CQs at the natural language layer \cite{Malheiros13}, but it also still has a 1:1 mapping and only three predefined types (isa, property value question (yes/no), and existence question). All three have been devised manually based on a manual analysis of CQs. As discussed and demonstrated in \cite{WPLK18}, 1:1 mappings are suboptimal, because a CQ may be represented in different ways in an OMS. For instance, a verb in a CQ need not be an object property in an OWL ontology, nor will a noun in a CQ necessarily be a class in the ontology, and both verbs and nouns will be nouns in thesauri; e.g., `marriage' vs. `married to', `advertising' vs `advertise' etc.. A difficulty with CQs is that they may require different formalisations to query the OMS and also even for just ontologies already, depending on the usage scenario: type-level queries would be formalised with, say, SPARQL-OWL \cite{Kollia11}, instance-level queries would map to SPARQL \cite{sparql}, and yet others relate to tests and presuppositions for axioms so may be formalised in, e.g., OWL \cite{Dennis17,Fernandez18,Malheiros13}. 

Given that a CNL for CQs is supposed to function for specifying requirements for any ontology, the logic-based knowledge representation must be decoupled from the natural language. At the same time, it is well-known that the other extreme---free-form sentences---makes it exceedingly hard to formalise, be this for query or axiom generation; e.g., most recently, Salgueiro et al.'s system allows free-text as input, but only four types of questions may generate answers in their IR-based approach (some definition questions, yes/no, facts, and lists) \cite{Salgueiro18}. A middle way to bridge this gap is to design a CNL. 

CNLs for computation have been proposed as a solution for various information management aspects, such as query formulation to hide SPARQL syntax (e.g., Sparklis \cite{Ferre17} and Quelo \cite{Franconi10}), generation of pseudo-NL sentences from axioms in an ontology to formalise them (e.g., ACE \cite{Fuchs10}), and software requirements formulation with, notably, the Semantics of Business Vocabulary and Rules (SBVR) \cite{sbvr08}. Recent literature reviews on CNLs within the scope of the Semantic Web can be found in \cite{BouayadAgha14,Safwat16} and more broadly on CNLs in \cite{kuhn2014cl}. They all---22 tools and proposals in \cite{Safwat16} and 22 in \cite{BouayadAgha14}---focus on assertions for ontology authoring, even those for queries, such as ``give me all writers who ...'' rather than ``which writers...?'', and even where they are questions, they are for instances, rather than the TBox-level of typical CQs, hence, take a different form. 

 Thus, to the best of our knowledge, there is no CNL for CQs for ontologies that is implementation-independent.

\section{CNL design}
\label{sec:cnl}

The overarching approach to the design of the CNL for CQs is a semi-automated and data-driven bottom-up approach. The data is taken from the novel dataset of CQ patterns of \cite{WPLK18} that were created automatically from a set of 234 quality CQs from 5 ontologies, which were analysed on their linguistic structures; this is summarised in Section~\ref{sec:cqpatterncreation}, so as to keep the paper self-contained. We analysed those CQ patterns, which informed the actual CNL design and specification that is described in Section~\ref{sec:design}.

\subsection{Preliminaries: CQ patterns}

The automated CQ pattern creation process by Wi\'sniewski et al. \cite{WPLK18} avails of 234 CQs that were collected from publicly available CQ sets for publicly available ontologies, which is the largest data set of CQs for ontologies. They were manually checked on whether they are TBox queries and at least questions, for it to merit inclusion in the dataset. Eventually, only five ontologies with their CQ sets passed these criteria, which were the CQs of the Software Ontology (SWO) \cite{Malone14}, the Dem@care \cite{DEMCARE} about care for patients with dementia, OntoDT \cite{Panov:2016:GOD:2869973.2870281} about data types, Stuff \cite{Keet14ekaw}, and African Wildlife (AWO) \cite{Keet18oebook} (refer to \cite{WPLK18} for further details on rationale and the CQ set).

These CQs were used to create {\em domain-independent CQ patterns}. A pattern here refers to the general structure of the question that is shared among more than CQ, be this for a single or multiple ontologies, and thus irrespective of an ontology's vocabulary.

\subsubsection{CQ pattern creation}
\label{sec:cqpatterncreation}

Wi\'sniewski et al. \cite{WPLK18} applied the following procedure for each CQ in their dataset of 234 CQs:

\begin{enumerate}
\item {\em Entity Chunk (EC) identification}: the EC is a fragment of text referring to an entity that is likely to be represented in an ontology. ECs contain nouns and noun phrases; 
e.g., in a CQ: {\sf\small What are the types of furry carnivorous animals?} both ``furry carnivorous animals'' and ``the types'' phrases should be identified. Yet, only ``furry carnivorous animals'' is likely to be represented in an ontology while ``types''  indicate the kind of element (a class, in this case). 
Therefore, a manual verification of ECs was performed to reject those ECs that are unlikely to have a mapping to an element in an ontology.
If an EC was accepted, a sequential number was added to separate the different ECs.
Thus, the CQ: {\sf\small What are the types of furry carnivorous animals?} would have been transformed into {\em What are the types of EC1?}.

\item {\em Predicate Chunk (PC) identification}: this is an analogous procedure to EC identification, but here the text fragments refer to entities that describe relations between elements that are likely to be represented in an ontology. The PCs are consecutive verbs and may contain adpositions/particles and they may have an auxiliary part that may be located in a different place of the question than the main part of PC. For instance, in the CQ {\sf\small What does this animal eat?} both ``eat'' and ``does'' are verbs, hence, they were classified as PCs, but ``does'' is an auxiliary verb to ``eat'', so the algorithm identified the whole phrase ``does eat'' as a single PC. 
As in the case of ECs, PCs were added with successive numbering; e.g., a {\sf\small Which country do I have to visit to see these animals?} would be chunked into {\em Which EC1 PC1 I PC1 to PC2 EC2?}, as ``do'' and ``have to visit'' belong together, and ``see'' is a second PC. 
There were PCs that would unlikely end up as relations/object properties in an ontology, such as 
``is'', ``are'', and ``have'', which has been manually checked so as to keep them as text chunk.

\item {\em Generalisable pattern selection}: The PC and EC identification generated a simplified, domain-independent, form of every CQ as ``candidate pattern''. 
To identify and extract the actual patterns from them, a distinction in treatment was made between what Wi\'sniewski et al. refer to as ``dematerialized'' and ``materialised'' CQs.
\begin{itemize}
\item Dematerialised CQ: the CQ has `replaceable' content already. For instance, SWO's {\sf\small What software can perform task x?}  is meant to be used such that the user fills in a real task from the ontology for the placeholder ``task x'', i.e., ``materialise'' the CQ. The CQs in the dataset have this indicated as a fragment of text in square brackets, like {\sf\small swo08.What software can perform [task x]?}. 
Therefore, even if such a dematerialised CQ was observed only once, by design it is intended for reuse, and thus it produced a CQ pattern.
\item Materialised CQ: each object is explicitly mentioned in the CQ. If a pattern candidate based on a materialised CQ was unique, i.e., 
it was rejected as CQ pattern. 
E.g., {\sf\small Is there an animal that does not drink water?}, or pattern candidate {\em Is there EC1 that PC1 EC2?}, did not repeat and is thus not in Wi\'sniewski et al.'s CQ set\footnote{Note that not all is lost: such a CQ can easily be reworded into {\sf\footnotesize Which animal does not drink water?} and chunked into {\em Which EC1 PC1 EC2?}, which does re-occur and, as we shall see, turn into a template in \cqcnl.}.
\end{itemize}
\end{enumerate}

\noindent 
Applying the EC and PC chunking extraction and selecting only those forms that are generalizable, resulted in a list of 106 domain-independent patterns out of the original 234 domain-specific CQs.

\subsubsection{Analysis of the CQ patterns}
\label{sec:analysePatterns}
We analyse these 106 patterns on both structural features, such as maximum number of variables and chunks in a sentence, and sentences' and patterns' meanings, such as the use of synonyms, singular/plural, and other aspects that may emerge on closer manual analysis of the patterns. These observations will then inform the specification of the CNL.

We first considered the anatomy of the CQ patterns. Text chunks can appear anywhere in the pattern, with a minimum of 0 text chunks and a maximum of 4 text chunks. Each pattern has at most 4 EC variables and 2 PC variables, and an overall of 5 variables in a pattern. Because a PC variable can be split up into different chunks (e.g., a ``do we need'' is chunked as  {\em PC1 we PC1}), the highest number of slots for the variables is 6. Split PCs either have another variable, text, or a single space between the slots, and there are at most 3 chunks for a PC variable.

					

Concerning variations in patterns, there are commonly known sources of variation, like synonyms, and others. Illustrative examples for each type of variation are as follows.
\begin{enumerate}
\item Singular/plural. Consider pattern {\em 82.What type of EC1 is EC2?} and {\em 83.What types of EC1 are EC2?}, which could be normalised to the singular form, if desired, and therewith facilitating the custom practice to use terms in the singular in ontologies.
\item Superfluous words in the sentence. For instance, the ``or not'' in  {\em 27.Is EC1 EC2 or not?} is redundant, as is ``possible'' in {\em 51.What are the possible types of EC1?}.
\item Impersonal and personal sentences and, therewith, patterns. A CQ alike {\sf\small swo37.Can we collaborate with developers of [software x]?} could also have been written as, say, ``Is it possible to collaborate with developers of [software x]?''. 
\item Synonym usage in the text chunks. Because of the restricted domain, there are not many options, but they do exist; e.g., ``kind of'' and ``type of'' are used synonymously in CQs, yet result in different patterns; e.g., {\em 79.What kind of EC1 is EC2?} and {\em 82.What type of EC1 is EC2?}.
\item The same information request can be formulated in different ways, such that it would need/use a different pattern. For instance, the CQ 
{\sf\small swo15.What software can I use [my data] with to support [my task]?} can be rewritten as, e.g., 
``Which software can use [my data] to support [my task]?'' as well as ``Which software can support [my task] with [my data]?'', which would result into the following two patterns:  
{\em What EC1 PC1 I PC1 EC2 PC2 EC3?}, 
{\em Which EC1 PC1 EC2 PC2 EC3?}, and
{\em Which EC1 PC1 EC2 with EC3?} 
respectively. 
\end{enumerate}

Noteworthy is that ``we'' and ``I'' only appear in the CQ set of the SWO that was created by a set of authors, ``kind of'' appears only in the AWO and Stuff CQ sets that were authored by the same author, ``type of'' appears only in the SWO CQ set, and ``types of'' appears only in the Dem@Care CQ set (exact author(s) unknown). That is, there seems to be either author preference or some (un)conscious authoring choice to generate more questions in the same way. This is not to say one way of formulating a CQ is better than another, merely to observe that different CQ sets seem to exhibit different sentence `styles' at least for a subset of their CQs.

Finally, negation---in the sense of both disjointness among classes and for a class' properties---is present in the CQs, but only once each and thus did not result in a pattern in \cite{WPLK18}. Ren et al.'s and Bezerra et al's Pizza example CQs for their templates with negation do not appear in the Pizza QC set, however, 
so even if that set would have been included in \cite{WPLK18}'s dataset, it would not have made a difference in the set of patterns detected by Wi\'sniewski et al.'s algorithms. The Pizza CQ set does have an imperative ``Find all pizzas that have prawns but not anchovy.'' In question format, this would be {\sf Which pizzas have prawns, but no anchovy?}, which can be chunked as {\em Which EC1 have EC2, but no EC3?}. If a more precise verb than ``have'' would be used, alike in aforementioned awo\_9 in the same question format ({\sf which animals do no drink water?}), then it chunks as {\em Which EC1 PC1 EC2?}, which also fits with Ren et al.'s ``Which pizza contains no mushroom?" \cite{Ren14} when reformulated as `which pizza does not contain mushroom'. The issue is analogous for the disjointness examples. 

\subsection{The \cqcnl~CNL}
\label{sec:design}

\subsubsection{Design considerations}

There are two extreme design options for a CNL, which is often template-based: 1) minimalist with the fewest amount of templates that are shortest and 2) including variants to allow flexibility and have better flowing text. The second option tends to be favoured when text has to be generated from structured data or knowledge so as to make the text not look `boring' (rigid and stale), whereas the first option is more prevalent in CNLs for ontologies (authoring and reading) and conceptual data model design, for it suggests that it would make the step toward model and axiom generation less hard. However, it is easy to extend the principles to specify multiple surface variants for one type of axiom or query as long as they are linked or recorded that they are variants of a `standard' or `default' template, rather than have the template structure adhere rigidly to the structure of a particular type of axiom. This approach has been proposed before for a temporal logic for temporal conceptual modelling \cite{Keet17creol}, which was based on a user evaluation on template preferences.

Because different authoring preferences or customs were detected in the dataset, we will keep all CQ patterns and convert them into templates, but also generate a `default' CQ template, where applicable. Because there are not that many CQ patterns, and therewith also unlikely to be many variants, a template-based approach will be taken for the CNL at this stage, rather than specification of a grammar. 

As last design consideration, while there is no negation in any of data-driven patterns of \cite{WPLK18}, there is in the CQ set and elsewhere; therefore, we deem it reasonable to add a few templates to cover these cases. Even though that hiding the negation makes it less cumbersome for a CNL, it will make it harder for processing it automatically into a query over the resource, whereas it is an easy signal in a template. 

\subsubsection{Specification}
The generation of the `default' templates applies to those CQ patterns of \cite{WPLK18} where there were issues or commonalities regarding, mainly: 1) singular/plural forms, 2) the I/we designations in a pattern, 3) removing redundant words in text chunks, and 4) synonym usage. 
To illustrate these changes, consider CQ pattern {\em 1.Are there any EC1 for EC2?}: it is in the plural and has the redundant ``any'' word, which therefore results in a template of {\em Is there [EC1] for [EC2]?}, which turned out to be identical to CQ pattern 30, and thus removed so as to obtain a list of unique sentences. 
CQ pattern {\em 67.What EC1 PC1 I PC1 EC2 on EC3?} and similar ones with ``I''/``we'' are harmonised into removing ``I''/``we'' and one of the PC1s, resulting in the template {\em 67.What [EC1] [PC1] [EC2] on [EC3]?}, which is reasonable given the original CQ {\sf\small swo86.What compiler do I need to compile source code on [platform x]?} which can be rewritten equivalently into {\sf\small What compiler is needed to compile source code on [platform x]?}\footnote{arguably, it is `which', not `what' for this particular CQ.}, i.e., the ``do I need'' and its corresponding pattern fragment {\em PC1 I PC1} can be captured with an ``is needed'' and pattern fragment {\em PC1}, hence the {\em PC1 I} is removed from the CQ pattern to generate the default template.

Regarding synonyms, `type of' was selected over `kind' and `category' for the defaults. This resulted in the merger of, e.g., CQ pattern {\em 79.What kind of EC1 is EC2?} and pattern 82 into template {\em 70.What type of [EC1] is [EC2]?} and of {\em  48.What are the main categories of EC1?} and pattern 49 into template {\em 42.What are the main types of [EC1]?}.

Applying these transformations manually throughout the list of 106 CQ patterns and removing any duplicates that were generated during this process resulted in 89 templates with 40 variants, where the variants have an additional letter designation; e.g., 
{\em 22.Is [EC1] [EC2]?} and variant 
{\em 22a.Is [EC1] [EC2] or not?}. 
The complete list of templates is included in the Appendix. 

Fourteen of the 89 templates are fragments of others; e.g., 
{\em 22.Is [EC1] [EC2]?} is a template fragment of  
{\em 23.Is [EC1] [EC2] for [EC3]} and  {\em 24.Is [EC1] [EC2] or [EC3]}. This may be of interest to further reduce the number of templates as well as be of interest for a predictive editor in tool design.

To cater for the negations, three basic templates were attached, so as to cover the cases of `does not PCi', `PCi no ECi', and class disjointness (numbers 90-92).

\subsubsection{Storing templates and CQs} 
\begin{figure}[h]
\centering
\includegraphics[width=0.9\textwidth]{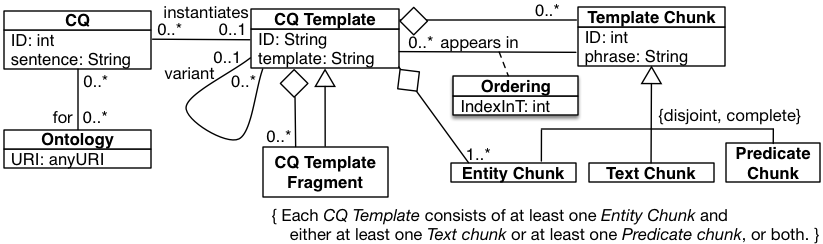} 
\caption{Data model for CQ templates.}\label{fig:claromodel}
\end{figure}

While CQ templates can be stored in a simple txt file, it serves to store them in a structured way so that multiple tools can use and analyse them in the same manner. To the best of our knowledge, there is no standard for storing a CNL. Therefore, we designed our own data model for storing CQ templates,  which is depicted in Fig.~\ref{fig:claromodel} in UML Class Diagram notation. The constraints represented in the model about {\sf CQ Template} do not violate the \cqcnl~data. We did consider whether there could be future CQ templates that would not have an {\sf Entity Chunk}, but even statements about processes will have them categorised as an {\sf Entity Chunk}, as in, e.g., ``is chewing involved in eating?'' (the processes are reified in the sentence).

In addition, to permit extensions to \cqcnl, there may be CQs that do not instantiate a template (hence, the {\sf 0..*} on {\sf instantiates}). Also, users should be permitted to author CQs without an ontology being present already, as this activity may happen before ontology development; hence, the {\sf 0..*} on the {\sf for} association end.

\section{Evaluation}
\label{sec:eval}

We conduct a preliminary evaluation of \cqcnl~to answer the following two questions:
\begin{enumerate}
\item[RQ1:] Does \cqcnl~cover the CQs from the training set? 
\item[RQ2:] Is \cqcnl~sufficiently comprehensive for unseen CQs?
\end{enumerate}
\cqcnl~should be able to deal with the CQs of the data set of \cite{WPLK18}, but may not, because not all CQs resulted in a pattern (recall that a ``pattern candidate'' (a chunked CQ) only became a pattern if it occurred more than once in the dataset of CQs \cite{WPLK18}). Also, the CQ patterns were obtained automatically and a verification was not performed on the CQ patterns. In addition, for the time being that there is no advanced CQ tool,  authors will author a question manually and thus may need to do the chunking themselves\footnote{Eventually, it should possible to run the algorithm on any new CQ fed into such a CQ authoring tool and automatically checked agains the extant CQ patterns and templates.}. 

The second question aims to assess whether \cqcnl~provides a broad enough coverage of possible sentence templates to be adequate beyond the training data.

Finally, we compare \cqcnl~to the templates of Ren et al. and Bezerra et al..

\subsection{Design}

\subsubsection{Methods}
To answer Question 1, we take a random selection of 10\% of the  CQs in the dataset and test them on authorability of the CNL constructed. This set is called SetA. Each sentence is manually chunked into ECs and PCs by one of the authors 
and then checked against \cqcnl's templates. For each CQ in SetA, record whether it can be authored in the CNL and, if not, why not, then compute percent coverage. Afterward, the manual chunking was compared against the  mapping of CQs to CQ patterns as well as to \cqcnl's templates that was kindly provided by D. Wi\'sniewski.

To answer Question 2, we collect a new set of CQs that are at least for a different ontology, that are authored by people other than those who authored the CQs in the data set, and are ideally also in a different domain. The target is 20 TBox-level (i.e., not instance-level) CQs. This set is called SetB. A second test set, SetC, is created from half of the Pizza ontology CQs  so that is amount to about the same size as SetB; they are kept separate, as there is some overlap in CQ authors  of the SWO and Pizza CQs.  
 For each CQ in SetB and SetC, record whether it can be authored in the CNL. If it cannot be authored directly, attempt to manually reformulate it into a sentence with equivalent meaning that does fit with one of the templates. Compute percent coverage for both the original set and the set with reformulations (if any). Compare the outcomes of SetA, SetB, and SetC.
 
The comparison with Ren et al. and Bezerra et al.'s templates is two-fold. First, we compare their respective templates to the \cqcnl~templates, with the alignment that their CE maps to \cqcnl's EC and their OP/OPE/DP to \cqcnl's PC. 
Second, from this comparison follows at least part, if not fully, the coverage of their template sets for the CQs in SetA, SetB, and SetC. If there is no equivalent template, then Ren et al.'s, respectively, Bezerra et al.'s template, is checked against the CQ in question, and tested against the CQs for which \cqcnl~does not have a fitting template (if applicable).  

\subsubsection{Materials}

For SetA, we take every 10th CQ from the list of \cite{WPLK18}, being: swo01, swo11, ... swo81, stuff\_03, awo\_2, awo\_12, DemCare\_CQ\_9, ... , DemCare\_CQ\_99, ontodt\_02, and ontodt\_12, resulting in a set of 24 CQs.

For SetB, we assess CQs from two recent and one related paper, starting with those described in \cite{Moreira18,Zemmouchi13} and filling it up to 20 with the CQ set of the Vicinity project\footnote{\url{http://vicinity.iot.linkeddata.es/vicinity/}; last accessed: 20 Dec. 2018.} that is being used for ontology testing \cite{Fernandez18}. The scopes of the ontologies that the CQs relate to are at least partially different from those in the dataset and, to the best of our knowledge, there is no overlap in CQ authors. The CQs are definitely different from those in Wi\'sniewski et al. \cite{WPLK18}'s CQ set.
The Pizza ontology CQs are sourced from R. Stevens' lecture slides\footnote{page 4 of \url{http://studentnet.cs.manchester.ac.uk/pgt/2014/COMP60421/slides/Week2-CQ.pdf}; last accessed: 9-1-2019.}. To have about equal number of CQs as in SetB, every other CQ in the list is selected, resulting in 21 sentences.

The CQ templates of Bezerra et al. and Ren et al. are taken as published in \cite{Bezerra14} and \cite{Ren14}, respectively.

All data and results are available as supplementary material at \url{https://github.com/mkeet/CLaRO}.

\subsection{Results and Discussion}
\label{sec:results}

\paragraph*{Verification with training set CQs}
Manual chunking of the CQs in SetA and testing against \cqcnl~yielded a 70.8\% initial success rate. Further analysis on alternative ways of chunking the sentences with the template list in hand, this increased to 83.3\% (20 out of 24). The four remaining cases are interesting and clearly demonstrate challenges with bottom-up approaches to designing a CNL. We chunked {\sf\small swo11.Which visualisation software is there for [this data] and what will \underline{it} cost?} as {\tt Which [visualisation software]$_{EC1}$ is there for [this data]$_{EC2}$ and what [will]$_{PC1}$ [\underline{it}]$_{EC1}$ [cost]$_{PC1}$?}, because of the referring expression ``it'' at the end of the sentence, whereas the automated chunker had not picked it up and had generated the CQ pattern {\em Which EC1 is there for EC2 and what PC1 \underline{EC3} PC1} that resulted in \cqcnl's template {\em Which [EC1] is there for [EC2] and what [PC1] [EC3] [PC1]?} If the template would have had just an `EC slot', rather than an enumerated EC slot, then it would have passed. 
Another CQ that failed was {\sf\small DemCare\_CQ\_99.What types of descriptive information are relevant to an observation?}, which is due to ambiguity: ``descriptive information'' could be redundant words to safely ignore, it could be a separate EC, or or could be grouped together with ``observation''. {\sf\small DemCare\_CQ\_29.Which are the tasks of the semi-directed step?} was initially chunked manually into {\tt Which are [the tasks]$_{EC1}$ of [the semi-directed step]$_{EC2}$?}, which does not have a matching template, nor did it have a CQ pattern in the original dataset because it was unique. However, grammatically, the CQ sentence should be `What...' not `Which...'. Correcting the sentence, the chunking changes into {\tt What are ...}, which is also not in \cqcnl, but the singular does have a matching template: {\em 60.What is [EC1] of [EC2]?}.  

The automatically chunked CQs had a 91.7\% initial success rate, and 100\% upon further analysis. The two that failed initially, DemCare\_CQ\_29 and DemCare\_CQ\_89, were sentences with unique sentence structures in the CQ set, and therefore did not qualify to become a CQ pattern, hence, did not enter \cqcnl~as such. Interestingly, the manual chunking of DemCare\_CQ\_89---different from the way the algorithm had done it---did match template 42. Similarly, {\sf\small swo71.Does [it] have a tutorial?} was manually chunked into 
{\tt [Does]$_{PC1}$ [it]$_{EC1}$ [have]$_{PC1}$ [a tutorial]$_{EC2}$?}
that matched \cqcnl's template 29, yet it had been chunked automatically into the CQ pattern {\em Does EC1 have EC2}, matching template 8.

Overall, the less-than-full coverage is largely due to manual vs. automated chunking in order to fill in a template, neither of which is a trivial activity. RQ1 can thus be answered in the affirmative, but noting the challenges to chunk it in the `right' way.

\paragraph*{Coverage of \cqcnl}

The results for SetB are mixed. Upfront already, five out of the 20 questions collected (25\%) were not CQs for ontologies, such as {\sf\small hero5.Why universities are organized into departments?} and {\sf\small saref7.How to represent tri-axial acceleration data from accelerometers of an ECG device?}: the former is an explainer questions and the latter asks for procedural information, neither of which apply to declarative information represented in ontologies. Of the remaining 15, five had a direct match with a \cqcnl~template. Three more matched after rewording the plural into the singular ({\sf\small What are ...} into {\sf\small What is ...}) and a rephrasing of {\sf\small vic6.Which are the relationships a partnership is involved in?} into the grammatically better {\sf\small Which relationships are involved in a partnership?}. Given that the `What are/is' chunking also appeared in SetA, hence, twice now, we add the following template as variant to \cqcnl: {\em 60a.What are [EC1] of [EC2]?}. 

The remaining seven did not have a match, of which six would match a template of {\em What is [EC1]?}, such as {\sf\small vic1.What is an organization?}. Such simple definition request CQs appeared only once as a materialised CQ in the original dataset (DemCare\_CQ\_4), and thus had not made it to pattern status. The remaining failing CQ {\sf\small hero3.What average size and duration have governing board?} and, in better English, ``What is the average size and duration of the governing board'', so chunked as
{\tt What is [the average size]$_{EC1}$ and [duration]$_{EC2}$ of [the governing board]$_{EC3}$?}
is an extended version of template number 60. Also, this CQ is actually two questions wrapped into one: ``What is the average size of the governing board?'' and ``What is the average duration of the governing board?'', which can be chunked into
{\tt What is [the average size/duration]$_{EC1}$ of [the governing board]$_{EC2}$?} 
and thus matches \cqcnl~template 60. 

When we extend \cqcnl~with the template {\em 93.What is [EC1]?}, then the coverage for SetB is 93.3\% out of the 15 valid CQs, 
and splitting up the hero3 CQ, it reaches 100\% of the valid CQs. 

SetC's coverage could be expected to be higher, because of overlap of CQ authors. First, though, 9 of the 21 CQs turned out to be invalid: there were five imperatives (e.g., {\sf\small pizza8.Find all vegetarian pizzas.}), two ABox queries, one that the ontology cannot answer, and one is an extra-ontological modelling discussion question. Of the remaining 12, four were successfully matched in the first round and six more after rewording, reaching 83.3\% coverage; e.g., {\sf\small pizza14.Are toppings organic?} would require a template {\em Are [EC1] [EC2]?} but only the singular is present as  template 22. Rewriting the imperatives into questions, such as {\sf\small Which pizzas are vegetarian pizzas?} for pizza8, all five passed immediately (e.g., template 78a for the reworded pizza8). This brought up the score to a coverage of 88.2\% of the valid CQs in SetC. SetB and SetC combined with the \cqcnl~update then has a coverage of 90.6\% of the valid CQs (29 out of 32).

Overall, \cqcnl's 131 templates can process unseen CQs with a good level of coverage, thereby answering RQ2 in the positive. However, given that 34.1\% of the questions in SetB and SetC turned out not to be proper CQs for ontologies and the different levels of coverage for SetB and SetC, this evaluation has to be considered preliminary. On the positive side, the percentage of improper CQs suggests that a CNL for CQs may be a welcome addition, so that CQ authors may be encouraged more to write grammatically better and answerable questions.

\paragraph*{Comparison of \cqcnl against related work}

Regarding Ren et al.'s 19 templates, one is not a question (R1b.``Find [CE1] with [CE2].''), two match after rewriting the template into grammatically correct English (from ``Be there ...'' into ``Is there...''), and two are ambiguous of which one does not have a match. The one that does not match, 
R12. ``Do [CE1] have [QM] values of [DP]?'', is  based on a CQ "Do pizzas have different values of size?, which is not in the Pizza QC set. The Pizza CQ set does have ``Do pizzas come in different sizes?'', which can be chunked into {\em PC1 EC1 PC1 EC2?}, which matches template number 29. 

Three of Bezerra et al's 14 templates do not have a matching \cqcnl template. The first one, B3	``From which + $<$property$>$ + $<$class$>$?'' is based on the sample sentence ``From which nation is American pizza?'', which is not in the Pizza CQ set and with that sample sentence, the template should have had two classes. 
The second mismatch is B9.``Are + $<$class$>$ + $<$class$>$disjoint?'' for which also a sample sentence was given that is not in the Pizza CQ set (``Are vegetarian pizza and non-vegetarian pizza disjoint?''). The idea of the question is the same as {\sf stuff\_04. Can a solution be a pure stuff?} of the training data and it can be reworded partially into the disjointness template 92 repeatedly. Last, B10.``	Which + $<$class$>$ + $<$property$>$ + $<$class$>$ + not + $<$property$>$ + $<$class$>$?	'', whose sample sentence is also not in the Pizza CQ set (``Which are the pizzas that have mozzarella topping but not have meat topping?''). Ignoring the awkward phrasing, the essence of the information request is an extension of R1h and \cqcnl's template 90 {\em Which EC1 does not PC1 EC2?}.

\begin{table}[t]
\caption{Summary of main results regarding coverage of the template sets; best values of the comparison are highlighted in italics.}\label{tab:inc}
\centering
\begin{tabular}{|p{2.5cm}|p{2cm}|c|c|c|c|}
\hline
& & \textbf{SetA} & \textbf{SetB} & \textbf{SetC} & \textbf{Combined}   \\
\hline \hline
 & $|$Total CQs$|$ &24 &20 &21 &65 \\ \cline{2-6}
   & $|$Valid CQs$|$ & 24 &15 &12 &51\\ \hline
\multirow{3}{*}{\textbf{Match}}    & Ren et al. &6 &5 &6 &17\\ \cline{2-6}
    &  Bezerra et al. &3 &3 &4 & 10\\ \cline{2-6}
     &  \cqcnl &\textit{20} &\textit{14} &\textit{11} &\textit{45} \\ \hline
\multirow{3}{2.5cm}{\textbf{Pct. coverage (valid CQs)}}  & Ren et al. &25 &33 &50 & 33\\ \cline{2-6}
        & Bezerra et al. & 13& 20& 33& 20\\ \cline{2-6}
         & \cqcnl & \textit{83} &\textit{93} &\textit{92} &\textit{88} \\ \hline
\end{tabular}
\end{table}

\section{CQ authoring tool}

We developed a tool to aid domain experts and CQ authors in writing questions so that they do not have to start from scratch. This sections details the tool design considerations, its main modules, and how each component achieves its function. 

\subsubsection{Design considerations}

\cqcnl's 134 existing templates cannot cover all grammatically well-structured and answerable CQs; hence, the tool should have a mechanism to accept new CQs. These new CQs may them be analysed in order to expand the coverage of the CNL. This flexibility is achieved by making the auto-complete functionality to only assist authors and not limit their input to the CNL's bounds. Moreover, since the CQs may be created also for artefacts similar to ontologies (e.g., thesauri) and ontologies not in OWL, we deemed it  best to create a stand-alone tool that is not tightly coupled with an existing KOS editor.

\begin{figure}[h]
\centering
\includegraphics[width=0.65\textwidth]{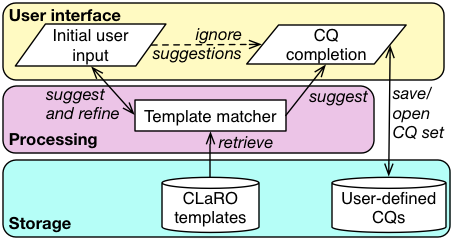}
\caption{Main components of the CQ authoring tool.}\label{fig:cqauthortoolrepr}
\end{figure}

\subsubsection{Components and implementation}

The main components of the tool are the user interface, template function module, and storage module as shown in Fig.~\ref{fig:cqauthortoolrepr}. The user interface is responsible for accepting the user's input, displaying user-friendly template suggestions, and listing all the user-defined CQs. The template function module is responsible for generating possible template suggestions given some user input and associating the final user input with a \cqcnl~template. The storage module is responsible for loading \cqcnl~templates from disk and loading/saving the user defined questions to disk. When saving the user-defined CQs to disk, the storage module serialises the set of user defined CQs according to an XML schema that has been developed based on the model described in Section~\ref{sec:design}.

\begin{figure}[t]
\centering
\includegraphics[width=0.9\textwidth]{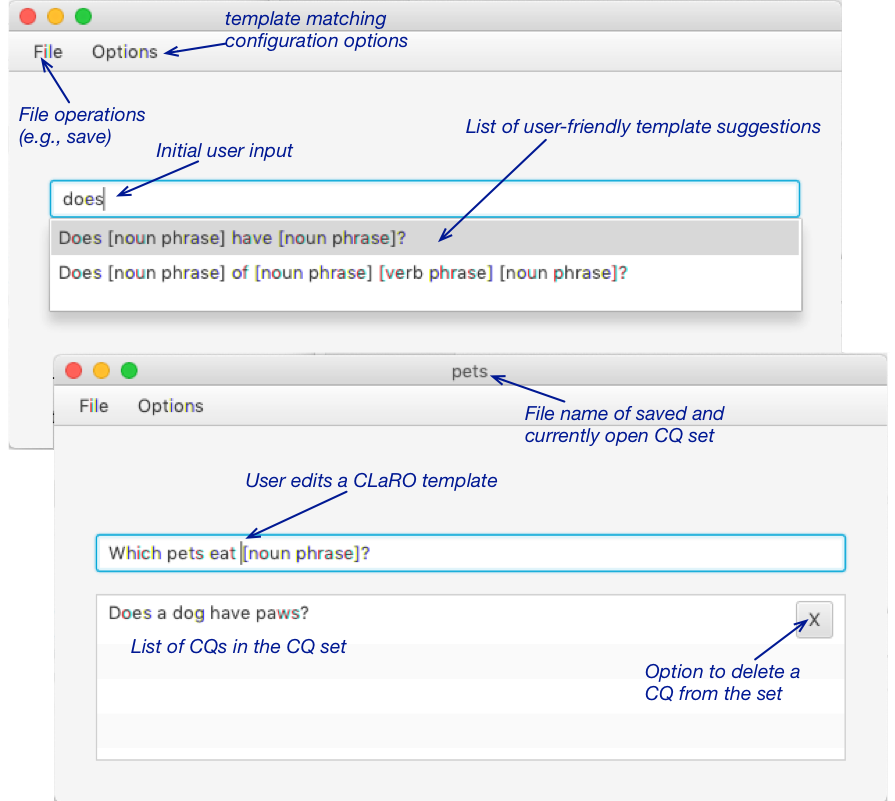}
\caption{Screenshots of the CQ authoring tool showing possible user input, autocomplete suggestions, file related actions, button for deleting a created question, and position of the filename where the user defined questions are stored.}\label{fig:cqauthortoolscreenshot}
\end{figure}

\subsubsection{Function and configuration}
When the user provides input through the interface, the system suggests a set of user-friendly forms of \cqcnl~templates as possible templates for a CQ. These user-friendly forms of \cqcnl~templates are generated within the autocomplete module by replacing all instances of the numbered abbreviations EC{\em i} and PC{\em i} for $i \in \mathbb{N}$ with the English full form ``noun phrase" and ``verb phrase" respectively from \cqcnl's templates. For instance, \cqcnl's template 1 is transformed from {\em Is there [EC1] for [EC2]?} to ``Is there [noun phrase] for [noun phrase]?''. The auto-complete function filters out non-relevant suggestions among the set of all possible ones for each some given instance of user input. A suggestion is considered relevant by the tool if it starts with or contains the user input. For instance, when the user types ``What type", then templates 70, 70a, and 71 are retrieved and rendered in their user-friendly form. The choice between the two types of relevance is configurable in the tool. Once the user selects a suggestion, they can edit the verb and noun phrase slots in order to obtain a question. They can also edit the selected template and write a question that does not fit within any \cqcnl~template. The templates and their corresponding \cqcnl~templates, if any, are then  saved to disk.

Two annotated screenshots of the tool are shown in Fig.~\ref{fig:cqauthortoolscreenshot}. In the example, when the user provides ``Does" as input, then the autocomplete feature will provide suggestions based on templates 8, 9, and 48. A suggestion based on template 48 is included only if the tool is configured to return all suggestions that contain the user input. The tool was not configured to return all templates with any match, hence, 48 is not included in this screenshot. Upon choosing the first suggestion, the user is able to create a CQ, as shown in the list of created questions in the unsaved document as indicated in screen 2 of Fig.~\ref{fig:cqauthortoolscreenshot}. In a single session, a user is able to create multiple questions and one can edit saved questions. Users do not necessarily have to create questions that adhere to the \cqcnl's templates since template suggestions can be ignored. For instance, the question ``Do Androids Dream of Electric Sheep?" does not adhere to any of \cqcnl's templates. Nonetheless, a user can create and add it to the list of stored CQs.

The current version of the tool is available as jar file together with its source code and screencast of its use are available at \url{https://
github.com/mkeet/CLaRO}.

\section{Examples and Use Case Scenarios}
\label{sec:ex}

In ontology, the typical uses for CQs include 1) informing the development process at the onset;  2) in the extension of an ontology as a maintenance option; 3) ontology verification, that is to ensure an ontology contains the expected contents (especially in large ontologies where there may be a need to have queries to be used in determining what knowledge it contains); and 4) for the evaluation of an ontology.  We illustrate several of these uses in the remainder of this section.

With \cqcnl's templates being fully domain independent, it is able to accommodate a CQ which is not covered by any of the CQ sets, encompassing typical uses 1 and 2, listed above. For instance, for an ontology about Italian pasta, a CQ {\sf\small Which method (do I need to use) to cook farfalle?}. This can be chunked as:\\
\indent {\tt Which [method]$_{EC1}$ [to cook]$_{PC1}$ [farfalle]$_{EC2}$?}\\ 
or, with the ``(do I need to use)'', it adds a, technically redundant, ``[PC1] I [PC1]'', and thus can be slotted into \cqcnl~template number 81.

A second example is illustrative for typical uses 2 and 3 or listed above as an overall pipeline, and is inspired by a CQ from the training set. Suppose a software user is interested in retrieving knowledge from the SWO or extending it. The user may wish to find out about what software can carry out a particular task, such as calculations or spellchecking, and thus may formulate the CQ {\sf\small What software can perform spelling correction?} (based on the CQ swo08) or, perhaps grammatically better  {\sf\small Which software can perform spelling correction?}. This CQ then either may be chunked automatically and matched against a \cqcnl~template, chunked manually by the author, or a template filled in in the \cqcnl~editor. Either way, the chunking would be, respectively:\\
\indent {\tt What [software]$_{EC1}$ [can perform]$_{PC1}$ [spelling correction]$_{EC2}$?}\\ 
\indent {\tt Which [software]$_{EC1}$ [can perform]$_{PC1}$ [spelling correction]$_{EC2}$?}\\ 
which corresponds to template number {\em 53.What [EC1] [PC1] [EC2]?} and to {\em 81.Which [EC1] [PC1] [EC2]?}, respectively.  Taking this further from finding a template to evaluating the CQ over the ontology, the data trail shows that template 53 matches Wi\'sniewski et al.'s dataset's CQ patterns 62 and 65.
The patterns, in turn, link to three SPARQL-OWL ``signatures'', which are templates for SPARQL-OWL queries that can be evaluated over an ontology when the terms from the CQ have been slotted in the place of the variables, including, among others the following query signature:

\begin{footnotesize}
\begin{verbatim}
SELECT  *
WHERE
  { ?placeholder_PPx1
            <http://www.w3.org/2000/01/rdf-schema#subClassOf>  _:b0 .
    _:b0    <http://www.w3.org/2002/07/owl#onProperty>  _:b1 ;
            <http://www.w3.org/2002/07/owl#someValuesFrom>  _:b2 .
    ?placeholder_PPx1
            <http://www.w3.org/2000/01/rdf-schema#subClassOf>  _:b3
  }
\end{verbatim}
\end{footnotesize}
where the ``{\tt \$PPx1\$}''  is a placeholder value, since swo08 has the variable {\sf\small [task x]} rather than the specific task spelling correction.

Illustrative of CQ authoring assistance, essential for smooth operation of typical sues 1 and 2 listed above is the following. 
29 of the 234 CQs of the dataset have the form of template 53 ({\em What...}), and 3 are of the form of template 81 ({\em Which...}); e.g.,  {\sf\small What data are measured for gait assessment?} (DemCare\_CQ\_40) that fits similarly with the template as\\  
\indent  {\tt What [data]$_{EC1}$ [are measured for]$_{PC1}$ [gait assessment]$_{EC2}$?}\\
Longer CQs tend to be more challenging. For instance, a CQ such as {\sf\small What is the set of datatypes that have [a datatype quality X] and [characterizing operation Y]?} (ontodt\_06), where ``{\sf\small [a datatype quality X]}''  is a placeholder for the value that a user wishes to query. In this case, it would be useful to have an advanced version of a CQ editor, that will not only assist the user to match the words to the variables in the templates, but also retrieve some of the knowledge from the ontology, so that the question posed will be answerable.
The CQ can be chunked into:\\
\indent {\tt What is [the set]$_{EC1}$ of [datatypes]$_{EC2}$ that have [a datatype}\\
\indent \indent {\tt quality X]$_{EC3}$ and [characterizing operation Y]$_{EC4}$?}\\
 and processed accordingly, i.e., being typed into template {\em  63.What is [EC1] of [EC2] that have [EC3] and [EC4]?}  or tagged and match to this template.
 
A different example of CQ authoring assistance is the notion of preventing authoring wrong ones. A use case of a CNL, and  \cqcnl~in particular, is that it could at least warn domain experts not to author `monstrous' CQs, like {\sf\small What is an ECG lead, what are the types of ECG leads, what type of property an ECG lead measures and what type of measurement an ECG lead can measure?} (saref3 in SetB, from \cite{Moreira18}). This lengthy CQ easily can be split up into 4 CQs without loss of knowledge ({\sf\small 	What is an ECG lead?}, {\sf\small What are the types of ECG leads?} etc.) and then it will fit with a template in \cqcnl~(in this case: 93 and 44, respectively). Likewise, it assists in preventing incorrect English, as was illustrated with CQ hero3 in Section~\ref{sec:results}.

Lastly on authoring assistance toward that notion of a pipeline to verification and evaluation, an example  concerning the flexibility in processing of a given CQ. 
As the sentences get longer, it becomes harder to chunk deterministically in one way only, questions may be formulated ambiguously, and one may argue about parsing rules and granularity. 
This may create difficulties for finding a matching template if there is no predictive editing, but also it may work in one's favour. For instance, take {\sf\small What is the set of datatype qualities for [a datatype X]?} (ontodt\_02), then it is conceivable to chunk it, or fill in the template's slots, in two different ways:\\
\indent {\tt What is [the set of datatype qualities]$_{EC1}$ for}\\
\indent\indent{\tt [a datatype X]$_{EC2}$?}\\
\indent {\tt What is [the set]$_{EC1}$ of [datatype qualities]$_{EC2}$ for}\\ 
\indent\indent {\tt [a datatype X]$_{EC3}$?}\\
In both cases, there is a \cqcnl~template: 38 and 61, respectively. Thus, even though there is only a limited list of CQ templates at present, it is already catering for flexibility in chunking and sentence variations. \cqcnl~does not help choosing which one of the two is the best option for a particular given situation, as it operates at the language layer, not the ontology layer.

\section{Discussion}
\label{sec:disc}

\cqcnl~is, to the best of our knowledge, the first CNL for competency questions for ontologies, surpassing the previously published archetypes and patterns \cite{Bezerra14,Malheiros13,Ren14} principally on the following aspects: i) decoupling of the language and cognition from the ontology artefact layer where design decision already have been taken, ii) larger number of types of questions supported, and iii) more variants in sentences structures to accommodate for several question formulation preferences.

Trying to find new CQs was a non-trivial endeavour, and of those we could find that were listed as CQs, it turned out that about a third of the questions were invalid as CQ. It is unclear what the main reason for that is, but it is certainly clear that \cqcnl~can assist with reducing that percentage for newly created CQs. Wi\'sniewski's et al.'s dataset \cite{WPLK18} does not have invalid CQs and they seem all grammatically correct (except, perhaps, an occasional and arguable `what'/`which'), which means they either have been curated upfront (not described to be the case in \cite{WPLK18}), or all the good CQ sets available went into that dataset, which is the more likely explanation. 

It was expected that the Pizza CQs (SetC) would yield a higher percentage of coverage than the other newly sourced CQs (SetB), due to the overlap in people involved in Pizza and SWO. This turned out to be the case in the strict sense: the original coverage for SetB before adding template 93 
to \cqcnl~was 53.3\% whereas for SetC it was 83.3\% for SetC. With the required manual interventions---a new template and rephrasing the imperatives---this increased the coverage to 93.3\% and 88.2\%, respectively, which is similar. That is, while good coverage can be obtained, it cannot be excluded that any possible intervention required for CQs from authors other than those who authored the CQs of the dataset, SetB, and SetC, may affect the set of templates in \cqcnl.

The model for storing the CQ template (Fig.~\ref{fig:claromodel}) may appear straightforward. To the best of our knowledge, however, there is no other model as precursor to an XML schema for storage of any CNL, even though there are template-based CNLs that are stored in XML notation. This model, therefore, may contribute toward the development of a {\em de facto} standard for storing template-based CNLs, not only for CQs, but generally for any CNL. This then may perhaps be wrapped in an extended version of, e.g., the NIF for NLP tool exchange of text annotations \cite{Hellmann13} when linked to a chunker for analysis free text CQs.

While the templates of \cqcnl~cover more sentence structures than the earlier proposed patterns and archetypes, the evaluation also has shown that more sentence structures may be possible than currently are covered with \cqcnl. Therefore, the \cqcnl~editor allows also new free-form CQs. A planned extension is to have the editor learn from the input given.

\section{Conclusions}
\label{sec:concl}

The paper presents the, to the best of our knowledge, first Controlled Natural Language for Competency Questions for ontologies: {\bf C}ompetency question {\bf La}nguage for specifying {\bf R}equirements for an {\bf O}ntology (\cqcnl). 
It was designed in a bottom-up way, availing of a new dataset of 234 competency questions that had been processed into 106 patterns. These patterns were analysed, and systematically converted into a template-based Controlled Natural Language, \cqcnl. The language was evaluated with questions from the training set and a small new set of competency questions, which demonstrated good to excellent coverage. Overall, the process resulted in 93 core templates and 41 variants, which cover over 90\% of the CQs of the test sets.

We are currently working on an intelligent editor for \cqcnl~in order to offer effective software-support for authoring competency questions.

\subsection*{Appendix: CQ templates}

The templates with an asterisk at the end were added after the evaluation.

\begin{multicols}{2}
\begin{footnotesize}
\noindent 
\textit{
1.Is there [EC1] for [EC2]? \\
1a.Are there any [EC1] for [EC2]?\\
2.Is there [EC1] [PC1] [EC2] [PC1]?\\
2a.Are there any [EC1] [PC1] [EC2] [PC1]?\\
3.Is there [EC1] to [EC2] [EC3] [PC1]?\\
3a.Are there any [EC1] to [EC2] [EC3] [PC1]?\\
4.Is there [EC1] in [EC2]?\\
4a.Are there [EC1] in [EC2]?\\
5.At what [EC1] [PC1] [EC2] of [EC3] [PC1]?\\
6.[PC1] [EC1] of [EC2]? \\
6a.Can we [PC1] [EC1] of [EC2]?\\
7.Who [PC1] [EC1] or [PC2] [EC2]?\\
7a.Do I know [EC1] who [PC1] [EC2] or\\
\mbox{ }\hspace{3mm} [PC1] [EC3]?\\
8.Does [EC1] have [EC2]?\\
9.Does [EC1] of [EC2] [PC1] [EC3]?\\
10.Given [EC1], what are [EC2] for [EC3] 
  of [EC4]?\\
11.Where [PC1] [EC1] [PC1]?\\
11a.How and where [PC1] [EC1] [PC1] 
  in the past?\\
12.How [EC1] is [EC2]?\\
13.How [EC1] [PC1] is [EC2] for [EC3]?\\
14.How long [PC1] [EC1] [PC1]?\\
15.How many [EC1] [PC1] [EC2]?\\
15a.How many [EC1] [PC1] I [PC1] [EC2]?\\
15b.How many [EC1] [PC1] we [PC1] 
  [EC2] [EC3]?\\
16.How [PC1] [EC1]?\\
16a.How [PC1] I [PC1] [EC1]?\\
17.How [PC1] [EC1] [PC1] [EC2]?\\
17a.How [PC1] I [PC1] [EC1] [PC1] [EC2]?\\
18.How [PC1] [EC1] with [EC2] [PC1]?\\
18a.How [PC1] I [PC1] [EC1] [PC1] [EC2]?\\
19.In what [EC1] [PC1] [EC2] [PC2]?\\
20.In what kind of [EC1] [PC1] 
  [EC2] [PC1]?\\
21.In which [EC1] are [EC2] in [EC3]?\\
22.Is [EC1] [EC2]?\\
22a.Is [EC1] [EC2] or not?\\
23.Is [EC1] [EC2] for [EC3]?\\
24.Is [EC1] [EC2] or [EC3]?\\
25.Is [EC1] of [EC2] [EC3]?\\
26.Is there [EC1] for [EC2] and where 
  [PC1] [EC3]?\\
26a.Is there any [EC1] for [EC2] and\\ 
\mbox{ }\hspace{3mm} where [PC1] I [PC1] [EC3]?\\
27.Is there [EC1] with [EC2]?\\
28.[PC1] [EC1] and [EC2] [PC1] [EC3]?\\
29.[PC1] [EC1] [PC1] [EC2]?\\
29a.[PC1] I [PC1] [EC1] [PC1] [EC2]?\\
29b.To what extent [PC1] [EC1] [PC1] 
  [EC2]?\\
30.[PC1] [EC1] [PC1] [EC2] that are 
  [EC3] from [EC4]?\\
31.[PC1] [EC1] [PC1] [EC2] to [EC3]?\\
32.[PC1] [EC1] [PC2] [EC2]?\\
33.[PC1] [EC1] if [EC2] [PC2] [EC3]?\\
33a.[PC1] I [PC1] [EC1] if [EC2] 
  [PC2] [EC3]?\\
34.[PC1] [EC1] on [EC2]?\\
34a.[PC1] I [PC1] [EC1] on [EC2]?\\
35.[PC1] some [EC1] of [EC2] for [EC3]?\\
35a.[PC1] I [PC1] some [EC1] of [EC2] 
  for [EC3]?\\
36.What are [EC1] and [EC2] for [EC3]?\\
37.What are [EC1] and [EC2] of [EC3]?\\
38.What is [EC1] for [EC2]?\\
38a.What are [EC1] for [EC2]?\\
39.What are [EC1] that have [EC2]?\\
40.What are [EC1] to [EC2]?\\
41.What are the differences between [EC1] 
  of [EC2]?\\
42.What are the main types of [EC1]?\\
42a.What are the main categories of [EC1]?\\
43.What are the main types of 
  [EC1] [EC2] [PC1]?\\
44.What are the types of [EC1]?\\
44a.What are the possible types of [EC1]?\\
45.What do [EC1] [PC1] [EC2] [EC3]?\\
46.What [EC1] are in [EC2] of [EC3]?\\
47.What [EC1] are of [EC2] with respect 
  to [EC3]?\\
48.What [EC1] does [EC2] have, 
  and what is its [EC3]?\\
49.What [EC1] from [EC2] [PC1] [EC3], 
  [EC4]?\\
50.What [EC1] is of [EC2] regarding [EC3]?\\
51.What [EC1] is of [EC2] regarding [EC3] 
  and [EC4]?\\
52.What [EC1] [PC1] [EC1] or [EC2] that 
  [PC2] [EC3]?\\
53.What [EC1] [PC1] [EC2]?\\
53a.What [EC1] [PC1] I [PC1] [EC2]?\\
54.What [EC1] [PC1] [EC2] given [EC3]?\\
55.What [EC1] [PC1] [EC2] [PC1]?\\
56.What [EC1] [PC1] [EC2] in [EC3]?\\
56a.What [EC1] [PC1] I [PC1] [EC2] 
  in [EC3]?\\
57.What [EC1] [PC1] [EC2] on [EC3]?\\
57a.What [EC1] [PC1] I [PC1] [EC2] 
  on [EC3]?\\
58.What [EC1] [PC1] [EC2] [PC1] [EC3]?\\
58a.What [EC1] [PC1] I [PC1] [EC2] 
  [PC1] [EC3]?\\
59.What [EC1] to [EC2] are there?\\
60.What is [EC1] of [EC2]?\\
60a.What are [EC1] of [EC2]?*\\
61.What is [EC1] of [EC2] for [EC3]?\\
62.What is [EC1] of [EC2] that have 
  [EC3]?\\
63.What is [EC1] of [EC2] that have 
  [EC3] and [EC4]?\\
64.What is [EC1] of [EC2] that have 
  [EC3] as [EC4]?\\
65.What is [EC1] of [EC2] that [PC1] 
  [EC3]?\\
66.What is [EC1] [PC1] [EC2]?\\
67.What is the difference between 
  [EC1] and [EC2]?\\
68.What [PC1] [EC1]?\\
69.What [PC1] [EC1] of [EC2]?\\
70.What type of [EC1] is [EC2]?\\
70a.What types of [EC1] are [EC2]?\\
70b.What kind of [EC1] is [EC2]?\\
71.What types of [EC1] [PC1] [EC1]?\\
72.When [PC1] [EC1] of [EC2] [PC1]?\\
73.Where do I categorise [EC1] like 
  [EC2]?\\
74.Where is [EC1] of [EC2]?\\
74a.Where's [EC1] of [EC2]?\\
75.Where [PC1] [EC1]?\\
75a.Where [PC1] I [PC1] [EC1]?\\
75b.Where [PC1] I [PC1] [EC1] [PC1]?\\
76.Where [PC1] [EC1] for [EC2]?\\
76a.Where [PC1] I [PC1] [EC1] for [EC2]?\\
77.Which are [EC1]?\\
78.Which [EC1] is [EC2]?\\
78a.Which [EC1] are [EC2]?\\
78b.What [EC1] are [EC2]?\\
78c.Which kind of [EC1] are [EC2]?\\
79.Which [EC1] is [EC2] of [EC3]?\\
79a.Which [EC1] are [EC2] of [EC3]?\\
80.Which [EC1] is there for [EC2] and \\
\mbox{ }\hspace{3mm}   what [PC1] [EC3] [PC1]?\\
81.Which [EC1] [PC1] [EC2]?\\
82.Which [EC1] [PC1] [EC2] [PC1]?\\
82a.Which [EC1] [PC1] I [PC1] [EC2] 
  [PC1]?\\
83.Which [EC1] [PC1] to [PC2] [EC2]?\\
83a.Which [EC1] [PC1] I [PC1] to [PC2] 
  [EC2]?\\
84.Which is [EC1] [PC1] [EC2]?\\
85.Which of the [EC1] and [EC2] [PC1] 
  [EC3] [PC1]?\\
86.Who is [EC1] of [EC2]?\\
86a.Who are [EC1] of [EC2]?\\
87.Who [PC1] [EC1] [EC2]?\\
87a.Who else [PC1] [EC1] [EC2]?\\
88.Who [PC1] [EC1]?\\
89.Who [PC1] [EC1] for [EC2]?\\
90.Which [EC1] does not [PC1] [EC2]?\\
91.What [EC1] [PC1] not [EC2]?\\
92.Which types are disjoint from [EC1]?\\
93.What is [EC1]?*}
\end{footnotesize}
\end{multicols}


\bibliographystyle{splncs03}
\bibliography{claro}

\end{document}